\documentclass[letterpaper, 10 pt, conference]{ieeeconf}  

\IEEEoverridecommandlockouts                              

\usepackage{amsmath,mathtools,array} 
\usepackage{amssymb}  
\usepackage{venndiagram}
\usepackage{nicematrix,tikz}
\usepackage{circuitikz}
\usepackage{algorithm}
\usepackage{float}

\newcolumntype{C}[1]{>{\centering\arraybackslash}m{#1}}

\usepackage{amsthm}\theoremstyle{definition}
\newtheorem{definition}{Definition}[section]

\title{\LARGE \bf
Causal discovery in deterministic discrete LTI-DAE systems
}

\author{Bala Rajesh Konkathi 
and Arun K. Tangirala
\thanks{$^{}$The authors are with the Department of Chemical Engineering,  
        Indian Institute of Technology Madras, Chennai, 600036, India.
        ({\tt\small email: balarajesh1729@gmail.com;  arunkt@iitm.ac.in})}%
}

\begin{document}

\maketitle
\thispagestyle{empty}
\pagestyle{empty}

\begin{abstract}
Discovering pure causes or driver variables in deterministic LTI systems is of vital importance in the data-driven reconstruction of causal networks. A recent work by Kathari and Tangirala in \cite{kathari2022novel}, 
formulated the causal discovery method as a constraint identification problem. The constraints are identified using a dynamic iterative PCA (DIPCA)-based approach for dynamical systems corrupted with Gaussian measurement errors. The DIPCA-based method works efficiently for dynamical systems devoid of any algebraic relations. However, several dynamical systems operate under feedback control and/or are coupled with conservation laws, leading to differential-algebraic (DAE) or mixed causal systems. In this work, a method, namely partition of variables (PoV), for causal discovery in LTI-DAE systems is proposed. This  method is superior to the method that was presented in \cite{kathari2022novel} as PoV also works for pure dynamical system which is devoid of algebraic equations. The proposed method identifies the causal drivers up to a minimal subset. PoV deploys DIPCA to first determine the number of algebraic relations ($n_a$), the number of dynamical relations ($n_d$) and the constraint matrix. Subsequently, the subsets are identified through an admissible partitioning of the constraint matrix 
 by finding the condition number of it. Case studies are presented to demonstrate the effectiveness of the proposed method.

 \end{abstract}

\section{Introduction} 
Causal analysis or discovery is a crucial task in many fields of study, including philosophy, economics, engineering and sciences, for the purposes of discovering the origins of phenomena, root-cause analysis, reconstruction of process topology, process network optimisation and disturbance propagation pathways. A problem of causal analysis involves discovering the directionality of influence between two or more variables, leading to the classification of variables into causes (drivers/sources) and effects (sinks). Mathematical treatment of the causal discovery problem necessitates a formal quantifiable definition of the ``causal'' relationship. Discovering such causal relationships from data entails using appropriate statistical inferencing/machine learning methodologies.\par

A few popular causality definitions include that of Granger, cross-mapping, intervention-based and, more recently, constraint-based causality. Each definition is constituted within a framework of assumptions and aimed at a certain class of processes. Granger causality (GC), arguably one of the most widely applied and tested definitions, for instance, rests on the axioms of temporal precedence (cause precedes effect), predictive potency of a cause and strictly within the walls of stochastic processes. While bivariate GC is devised for detecting \emph{total} causal influences, i.e., along \emph{direct} and \emph{mediated} or \emph{indirect} pathways, multivariate GC measures \emph{direct} influences. Cross-mapping, popularly known as convergent cross-mapping (CCM), is devised for sensing \emph{total} causal influences in weakly coupled nonlinear deterministic dynamical systems and primarily rests on the philosophy of inversion (as against prediction), i.e., ability to recover cause from effect. Extension to multivariable systems for detecting \emph{direct} causal influences using a multivariable state-space reconstruction method has been recently proposed by \cite{NithyaTangirala2021}. Positioned between these two classes of processes is the recently proposed \emph{dynamical constraint-based} form of causality for linear time-invariant (LTI) systems, which is the focus of this work.\par

The dynamical constraint-based causality as proposed in \cite{kathari2022novel} states that a signal $x_i[k]$ constraint-causes another signal $x_j[k]$ of an LTI dynamical system if they are tied together by a linear constraint of the form 
\begin{align}
    \mathbf{A}\mathbf{x}[k,L] = \mathbf{0}
\end{align}
where
\begin{align}
    \mathbf{x}[k,L] & = \left [\begin{matrix}
        x_i[k] & x_j[k] & x_i[k-1] & x_j[k-1] & \cdots \end{matrix} \right . \nonumber \\
        & \left . \begin{matrix} \quad \; \; x_i[k-L] & x_j[k-L] \end{matrix} \right ] ^T\label{eq:extvec}, \\
    \mathbf{A} & = \begin{bmatrix}
        0 & a_{1,j}^{(0)} & a_{1,i}^{(1)} & a_{1,j}^{(1)} & \cdots & a_{1,i}^{(L)} & a_{1,j}^{(L)}
    \end{bmatrix}
\end{align}
\noindent with the requirement that $a_{1,j}^{(0)} \neq 0$ and $a_{1,i}^{(l)} \neq 0$ at least for some $l \in [1,L]$. The first element of $\mathbf{A}$ 
should necessarily be zero for \emph{temporal precedence} to hold. For the multivariable case with $M$ variables and $n_d$ dynamical constraints, the extended vector in \eqref{eq:extvec} is constructed such that $\mathbf{x}[k,L] \in \mathbb{R}^{M(L+1) \times 1}$ so as to contain the instantaneous terms of all variables followed by their lagged versions. Accordingly, the constraint matrix is $\mathbf{A} \in \mathbb{R}^{n_d \times M(L+1)}$ with mandatory zeros corresponding to the instantaneous terms of all drivers (causal variables) in those constraints (refer to the Example given in Section \ref{sec:probstate}). The causal discovery problem is primarily that of identifying the so-called \emph{pure sources} or ``free variables'' from data without the knowledge of $\mathbf{A}$ and its rank $n_d$. The number of such pure sources is clearly $n_S = M - n_d$.
\begin{definition}
     A pure source is a variable that exerts influence on other variables without being influenced by any other. The cardinality of set of such sources in a system equals the degrees of freedom of system. 
\end{definition}

The work in \cite{kathari2022novel} presented a dynamic iterative PCA (DIPCA) - based algorithm to identify the number of pure sources $n_S$, $n_d$ and a basis for $\mathbf{A}$, i.e., $\mathbf{T}\mathbf{A}$, where $\mathbf{T} \in \mathbb{R}^{n_d \times n_d}$ is non-singular. A key result of relevance is that $\mathbf{A}$ has $n_S$ columns of zeros pertaining to the instantaneous terms of pure sources and that it remains invariant to $\mathbf{T}$. This pattern is the basis for the algorithm to identify both $n_S$ and the pure causal variables. An important limitation of such an approach is that the pattern breaks down when the variables are additionally bound by algebraic constraints, i.e., for the so-called (LTI-DAE) differential algebraic equation systems.\par

Causal discovery in LTI-DAE systems has not been attempted in the reported literature, although, as previously remarked, causality analysis of dynamical systems has been a subject of great interest. There has also been considerable study of causal discovery in algebraic or static systems; however, largely confined to systems with stochastic descriptions. DAE systems are not uncommon in engineering and other fields owing to the presence of inherent feedback, conservation constraints, etc. They are known to present challenges in other tasks such as simulation, identification, filtering and estimation.

This work aims to develop a method for identifying pure sources or causes in LTI-DAE systems from noisy data in the constraint-based causality definition framework. We propose a method that integrates DIPCA-based constraint identification with linear algebraic concepts. We term this integration the `Partition-of-Variables' (PoV) method. In addition, we demonstrate that the presence of algebraic equations precludes a unique identification of the pure sources but rather only allows the discovery of candidate sets of pure sources. This is attributed to the fact that the no specific pattern in $\mathbf{A}$ is preserved across its rotated versions, i.e., the constraints relating causes and effects can be written in different forms. The proposed algorithm identifies all such sets by partitioning of the identified constraint matrix into two parts $\mathbf{A} = \begin{bmatrix} \mathbf{A}_\text{dep} & \mathbf{A}_\text{r} \end{bmatrix}$, where $\mathbf{A}_\text{dep} \in \mathbb{R}^{(n_d + n_a) \times (n_d + n_a)}$ is the matrix corresponding to the instantaneous terms of $(n_d+n_a)$ admissible effects (or dependent variables) and by requiring that $\mathbf{A}_\text{dep}$ be of full rank. While the full rank requirement is useful for theoretical and noise-free cases, in noisy conditions, we replace this requirement with a threshold on the condition number of $\mathbf{A}_\text{dep}$. It is worthwhile noting that this method is applicable even when $n_a = 0$, thereby making it superior to the algorithm of \cite{kathari2022novel}.\par

The core idea in the proposed PoV-DIPCA approach is to solve the complementary problem of identifying the admissible dependent set of variables, i.e., set of admissible effect (sink) variables, in contrast to the approach of directly identifying the pure causal variables, followed in \cite{kathari2022novel}. This is achieved by using a rank condition on the partitioned constraint matrix corresponding to set of candidate-dependent variables. \par
 
The rest of the article is organised as follows. Section \ref{sec:probstate} presents 
the problem statement with a motivating example. Section \ref{sec:proposed_method} elucidates the proposed methodology
with the necessary theory and algorithm. Case studies are presented in Section \ref{sec:case_studies}. The paper ends with a few concluding remarks and directions for future work in Section \ref{sec:conclusions}.

\section{Background and Problem statement \label{sec:probstate}}

Consider an LTI-DAE system of $M$ variables $\mathbf{x}[k] = \begin{bmatrix} x_1[k] & x_2[k] & \cdots & x_M[k] \end{bmatrix}^T \in \mathbb{R}^{M \times 1}$ governed (constrained) by $n_d$ difference equations and $n_a$ algebraic equations, written as
\begin{align}
    \mathbf{A}\mathbf{x}[k,\eta] = \mathbf{0}, && \mathbf{A} \in \mathbb{R}^{(n_a+n_d) \times M(\eta+1)}, \label{eq:dae_sys}
\end{align}
where  $\mathbf{x}[k,\eta]$ contains both the instantaneous and lagged (up to $\eta$, the so-called process order) variables
\begin{align}
        \mathbf{x}[k,\eta] & = \begin{bmatrix} \mathbf{x}^i[k] \\ \mathbf{x}^l[k] \end{bmatrix}  \label{eq:xeta_xixl},\\
        \mathbf{x}^i[k] & = \mathbf{x}[k], \\
        \mathbf{x}^l[k] & = \begin{bmatrix} \mathbf{x}^T[k-1] & \mathbf{x}^T[k-2] & \cdots & \mathbf{x}^T[k-\eta] \end{bmatrix}^T, \label{eq:veclagvar}
\end{align}
and $M > n_d + n_a$, resulting in $n_S = M - (n_d + n_a)$ free variables/pure sources. Corresponding to the instantaneous and lagged components of $\mathbf{x}[k,{\eta}]$ in \eqref{eq:xeta_xixl}, we partition $\mathbf{A}$ of \eqref{eq:dae_sys} into $\mathbf{A}^i$ and $\mathbf{A}^l$, respectively, such that
\begin{align}
\begin{bmatrix}
    \mathbf{A}^i & \mathbf{A}^l \end{bmatrix}\begin{bmatrix}
    \mathbf{x}^i[k] \\ \mathbf{x}^l[k] \end{bmatrix} = \mathbf{0}, \\
\Longrightarrow \mathbf{A}^i\mathbf{x}^i[k] + \mathbf{A}^l\mathbf{x}^l[k] = \mathbf{0},  \\
\mathbf{x}^i[k] = \underbrace{-\mathbf{A}^{i\dag}\mathbf{A}^l}_{\mathbf{R}}\mathbf{x}^l[k] \label{eq:regression form}
\end{align}
where $\mathbf{A}^{i\dag}$ is the pseudo-inverse of full row matrix $\mathbf{A}^{i}$. In the absence of any algebraic constraints, the matrix $\mathbf{A}^i$ contains $n_S$ columns of zeros at indices corresponding to the pure source variables. Consequently, the regression matrix $\mathbf{R}$ in \eqref{eq:regression form} contains $n_S$ row of zeros corresponding to the row indices of pure sources. These patterns cannot be guaranteed for DAE systems, which is the challenge addressed in this work. The example below effectively illustrates these points.

\subsection{Motivational example}\label{Motivational example}

Consider the following four-variable system consisting of two pure sources $u_1[k]$ and $u_2[k]$,
\begin{subequations}
\label{eq:ex1}
	\begin{align}
	y_1[k] & = 0.5y_1[k-1] + 2u_1[k-1] \\
	y_2[k] & = 0.8y_2[k-1] + 1.2u_1[k-1] + u_2[k-1].
\end{align}
\end{subequations}
Assigning ${\mathbf{x}}[k] \equiv \begin{bmatrix} y_1[k] & y_2[k] & u_1[k] & u_2[k] \end{bmatrix}^T$, we have that $M = 4$, $n_d = 2$, $n_a = 0$, $\eta = 1$ and, 
\begin{align}
	{\mathbf{x}}[k,{\eta}] & = \begin{bmatrix}
	    \mathbf{x}^i[k]\\
     \mathbf{x}^l[k]
	\end{bmatrix} = \begin{bmatrix} {\mathbf{x}}[k] \\ {\mathbf{x}}[k-1]	\end{bmatrix},\\
{\mathbf{A}}
& = \begin{bmatrix} 1 & 0 & 0 & 0 & -0.5 & 0 & -2 & 0 \\ 0 & 1 & 0 & 0 & 0 & -0.8 & -1.2 & -1 \end{bmatrix}, \\
\mathbf{A}^i &= \begin{bmatrix}
    1&0&0&0\\
    0&1&0&0
\end{bmatrix}.
\end{align}
The pure sources are recognized by a column of zeros in $\mathbf{A}^i$, 
i.e., the coefficients of $\mathbf{x}^i[k]$ in ${\mathbf{A}}$. 
Notice that $\text{rank}(\mathbf{A})
= 2$. Re-writing \eqref{eq:ex1} in regression form of \eqref{eq:regression form}, 
\begin{align}
\mathbf{x}^i[k] = \mathbf{R}\mathbf{x}^{l}[k], && \mathbf{R} = \begin{bmatrix} 0.5 & 0 & 2 & 0 \\
         0 & 0.8 & 1.2 & 1 \\
         0 & 0 & 0 & 0 \\
         0 & 0 & 0 & 0 \\
    \end{bmatrix}
\end{align}
showing two rows of zeros corresponding to pure sources. 

Now, suppose that an additional algebraic constraint arises (e.g., due to feedback), 
\begin{align}
    u_1[k] = 2(r_1[k] - y_1[k]) \label{algebraic constraint - ex1}
\end{align}
where $r_1[k]$ is a reference signal. With this new constraint, $\mathbf{x}[k] = \begin{bmatrix} y_1[k] & y_2[k] & r_1[k] & u_1[k] & u_2[k] \end{bmatrix}^T$, $n_a = 1$, $n_S = 5 - (2 +1) = 2$ and $\mathbf{A}^i$ 
changes to
\begin{align}
    \mathbf{A}^i
    = \begin{bmatrix}
        1 & 0 & 0 & 0 & 0 \\ 0 & 1 & 0 & 0 & 0 \\ 2 & 0 & -2 & 1 & 0
    \end{bmatrix} \label{matrix A_i in the example}
\end{align}
 clearly showing only one column of zeros as against two. One of the pure sources is unambiguously $u_2[k]$. The challenge is in finding the other source, hidden in the algebraic equation.\par 
 
\subsection{Objective}
The objective of the present work is to determine the pure sources in the discrete LTI-DAE systems given the measurements of variables $x_i[k]$,  $i=1,\ldots, M$, assumed to be corrupted with zero-mean Gaussian white errors of variances $\sigma^2_{e_i}$,
\begin{align}
    z_i[k] =x_i[k]+e_i[k],\quad e_i[k]\sim \text{GWN}(0,\sigma_i^2).
\end{align}
Also, the error variances are realistically assumed to be heteroskedastic.
Note that $\eta$, $n_a$, $n_d$, $\sigma^2_{e_1}$, $\sigma^2_{e_2}$, \ldots $\sigma^2_{e_M}$ and number of pure sources ($n_S$) are all unknowns.\par

\section{Proposed methodology}\label{sec:proposed_method}
\subsection{Theoretical analysis}

In a system comprising $M$ variables and $n_a+n_d$ equations, where $M>n_a+n_d$, there exists $M-n_a-n_d$ free variables and $n_a+n_d$ dependent variables. The whole set of variables is partitioned into two subsets of cardinalities, $n_a+n_d$ and $M-n_a-n_d$, respectively. 
The coefficient matrix for only the dependent variables set, utilizing all equations within the system, attains full rank, which equals $n_a+n_d$. Hence, the dependent variables set naturally excludes the free variables. Using this principle, the pure source variables are determined from all the $\binom{M}{n_a+n_d}$ combinations among $M$ variables. In this work, the assumption is that at least a pure source exists in the system, satisfying $M-n_a-n_d>0$. \par

In the current analysis, our focus is solely on the \textit{instantaneous terms} of the set of $n_a + n_d$ variables in (\ref{eq:dae_sys}), to calculate the rank. Consequently, $\mathbf{x}_\text{dependent} \in \mathbb{R}^{(n_a+n_d)\times 1}$ is defined which encompasses the instantaneous terms of the dependent variables set. Similarly,  $\mathbf{x}_\text{rest} \in \mathbb{R}^{((M-n_a-n_d)+M\eta)\times 1}$ is introduced that includes both the instantaneous terms of the free variables and all lagged variables.

Writing (\ref{eq:dae_sys}) in the form
\begin{align}
\begin{bmatrix}
\mathbf{A}_{\text{dep}}&\mathbf{A}_{\text{r}} 
\end{bmatrix}\begin{bmatrix}
    \mathbf{x}_\text{dependent}\\\mathbf{x}_\text{rest}
\end{bmatrix}=\mathbf{0},\\
\mathbf{A}_{\text{dep}}\mathbf{x}_{\text{dependent}}+\mathbf{A}_{\text{r}}\mathbf{x}_{\text{rest}}=\mathbf{0},\\
\mathbf{A}_{\text{dep}}\mathbf{x}_{\text{dependent}}=-\mathbf{A}_{\text{r}}\mathbf{x}_{\text{rest}}\label{partition of variables},
\end{align}
and the rank of $\mathbf{A}_{\text{dep}}$ is evaluated,
where $\mathbf{A}_{\text{dep}}\in\mathbb{R}^{(n_a+n_d)\times(n_a+n_d)}$ and $\mathbf{A}_{\text{r}}\in \mathbb{R}^{(n_a+n_d)\times((M-n_a-n_d)+M\eta)}$. 
\par

To illustrate, from the extension of Section \ref{Motivational example} example with five variables $\{y_1, y_2, r_1, u_1, u_2\}$ and three equations ($n_a = 1$ and $n_d = 2$)  contains three dependent variables and two pure sources $(n_S = M - n_a - n_d = 5 - 1 - 2 = 2)$. Suppose, for a combination of probable dependent variables among the $\binom{5}{3}$ possibilities in this example, rewriting (\ref{eq:dae_sys}) as (\ref{partition of variables}),

\begin{align}
    & \mathbf{A}_\text{dep,1}\begin{bmatrix}
        y_1[k] \\ y_2[k] \\ u_1[k] \end{bmatrix} = -\mathbf{A}_\text{r,1}
        \begin{bmatrix}
            r_1[k] \\ u_2[k] \\ y_1[k-1] \\ y_2[k-1] \\ u_1[k-1] \\ r_1[k-1] \\ u_2[k-1]
        \end{bmatrix} \label{a combination in the example}\\ \label{ex1 - a combination}
        & \mathbf{A}_\text{dep,1} = \begin{bmatrix}
        1 & 0 & 0 \\ 0 & 1 & 0 \\ 2 & 0 & 1\end{bmatrix} 
\end{align}
where $\mathbf{A}_\text{dep,1}$ has a full rank. Here, the subscript ${dep}$ and $1$ indicate the probable dependent set of variables and a combination from $\binom{5}{3}$, respectively.

If the rank of $\mathbf{A}_{\text{dep}}$ is full, then that combination $\{y_1,y_2,u_1\}$ in LHS represents a dependent set of variables and the remaining variables (not the lagged ones) are the free variables. Therefore, $\{u_2,r_1\}$ are considered as pure sources.\par
 
 \subsubsection*{Variables that contribute to rank deficiency of $\mathbf{A}_\text{dep}$} Variables whose instantaneous terms with zero coefficients for an entire column in $\mathbf{A}$, are not part of any of the algebraic equations. Apparently, such variables in the $n_a+n_d$ dependent set of variables contribute to the rank deficiency of the coefficient matrix $\mathbf{A}_\text{dep}$. Therefore, these variables serve as \textit{unambiguous pure source} variables.\par

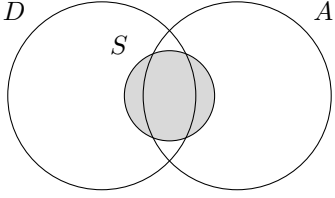
\begin{figure}[h]
     \centering
     \begin{tikzpicture}
  \node [draw,
    circle,
    minimum size =1.20cm,,fill=gray!30,
    label={135:$S$}] (S) at (0.9,0){};
\node [draw,
    circle,
    minimum size =2.5cm,
    label={135:$D$}] (D) at (0,0){};
 
\node [draw,
    circle,
    minimum size =2.5cm,
    label={45:$A$}] (A) at (1.8,0){}; 
\end{tikzpicture}
     \caption{$D$ - set of variables part of difference equations,
      $A$ - set of variables part of algebraic equations,
     $S$ - set of pure sources.}
     \label{fig: Euler diagram}
 \end{figure}

In conjunction with the sources that do form part of `sole' difference equations, as mentioned in the above paragraph, there may also be a few sources within the algebraic equations themselves. Particularly, not all algebraic equations may contain pure sources within the system as shown in the Euler diagram in Fig. \ref{fig: Euler diagram}.
\par

 Due to the presence of a source variable in the algebraic equation, the other variables in that equation are instantaneously related to the pure sources. They may also act as potential source candidates and, therefore, are referred to as \textit{ambiguous pure sources}. Consequently, $n_S$ \textit{appears} to be more than $M-n_a-n_d$ free variables. For free variables' determination, this appearance implies the need to compute the rank of \textit{all combinations}. 
 Only those combinations of variables that result in the full rank of $\mathbf{A}_\text{dep}$ are considered for pure sources determination and are termed \textit{admissible combinations}. Let choosing the combinations of variables for free variables determination be denoted as $\alpha$. If a combination is admissible, then $\alpha=1$; otherwise, $\alpha=0$.\par
 
From example \ref{Motivational example},
despite obtaining an exact number of sources ($n_S=2$) from a combination shown above in (\ref{a combination in the example}), we calculate the rank of $\mathbf{A}_\text{dep}$ for probable dependent variables across all $\binom{5}{3}$ combinations. The combinations are presented in Table \ref{Table - Example 1 combinations}. Notably, within the admissible combinations, when a variable is never considered as a dependent variable, it appears as an unambiguous pure source. Hence, the variable `$u_2$' appeared as an unambiguous pure source in the pair of admissible combinations. In contrast, the variables `$r_1$' or `$u_1$' appeared as pure source as they are related algebraically in (\ref{algebraic constraint - ex1}). Thus, they are potential candidates for other remaining source called ambiguous pure source.

\begin{table}[h!]
    \centering
    \caption{All combinations of 3 variables of instantaneous terms}
    \begin{tabular}{|>{\centering}p{2.4cm}|c|c|c|}
    \hline
    Candidate dependent variables & Rank of $\mathbf{A}_\text{dep}$& $\alpha$ & Free variable\\
    \hline
    $\{y_1[k], y_2[k],u_1[k]\}$ &3& 1& $\{u_2,r_1\}$\\
    $\{y_1[k],y_2[k],r_1[k]\}$ &3& 1& $\{u_2,u_1\}$\\
    $\{y_1[k],y_2[k],u_2[k]\}$ &2& 0& -\\
    $\{y_1[k],u_1[k],u_2[k]\}$ &2& 0& -\\
    $\{y_1[k],u_1[k],r_1[k]\}$ &2& 0& -\\
    $\{y_1[k],u_2[k],r_1[k]\}$ &2& 0& -\\
    $\{y_2[k],u_1[k],u_2[k]\}$&2& 0& -\\
    $\{y_2[k],u_1[k],r_1[k]\}$&2 &0& -\\
    $\{y_2[k],u_2[k],r_1[k]\}$&2& 0&-\\
    $\{u_1[k],u_2[k],r_1[k]\}$ &1& 0& -\\
    \hline
    \end{tabular}
    \label{Table - Example 1 combinations}
\end{table}
 
 \subsection{Data-driven approach - causal analysis from the data}
 In reality, only the measurements of the variables are accessible but not the model equations. Using the framework of DIPCA on these measurements,  
 the number of types of constraints (algebraic and dynamical) is estimated, and the inferences of causality are made. \par
 
With $N$ observations of each variable, the data matrix is constructed as
\begin{align}
    \mathbf{Z} = \frac{1}{\sqrt{N}}\begin{bmatrix}
        \mathbf{z}_{1,n}\ldots \mathbf{z}_{M,n}&\mathbf{z}_{1,n-1}\ldots \mathbf{z}_{M,n-1}\ldots&\mathbf{z}_{M,n-L}
    \end{bmatrix},\\
    \mathbf{z}_{1,n}=\begin{bmatrix}
        z_1[n] &z_1[n+1]&\ldots z_1[n+N-1]
    \end{bmatrix}^T.
\end{align} 
 DIPCA on the noise-free measurements of the variables results in zero-valued singular values when relationships exist among the variables. The number of zero-valued singular values equals the number of relationships. However, when 
applied to noisy measurements of the variables, it fails to reveal the zero-valued singular values associated with the relationships among the variables. To display the existence of relations and the relations, the variables are scaled with their respective error standard deviations, mapping zero-valued singular values to unity singular values \cite{narasimhan2008model}. However, the error variances of the variables are unknown.\par
 
  It has been discussed in \cite{VishweshRamanathan} to determine the order ($\eta$), number of algebraic constraints ($n_a$), and number of dynamical constraints ($n_d$) with an assumption that the order is the same for all the difference equations. 
  Reviewing here the process of determining $\eta$, $n_a$, and $n_d$ by relaxing this assumption of the same order for all the difference equations. If $d_L$ is the number of all the linear constraints that relate the lag ($L$) used for constructing the data samples among the variables, then 
 \begin{align}
     d_L&=(L+1)n_a+(L-\eta_1+1)n_{d_1}+\ldots(L-\eta_p+1)n_{d_p}\label{equation relating lag, order and number of equations},\\
     n_d&=n_{d_1}+n_{d_2}+n_{d_3}+\ldots+n_{d_p} \label{sum of number of dynamical equations}.
 \end{align}
 The first term of RHS in (\ref{equation relating lag, order and number of equations}) refers to the number of algebraic equations corresponding to each lag $L$ used, the second term of RHS corresponds to the number of difference equations ($n_{d_1}$) of order $\eta_1$ and likewise for the rest of the terms. $\eta_p$ is the maximum order of the system with corresponding $n_{d_p}$ difference equations. The unknowns 
 $n_a$, $n_{d_1}$, $\eta_1$, $n_{d_2}$, $\eta_2$ $\ldots$ and $n_d$ can be estimated using DIPCA of the measurements. However, it is sufficient to estimate $\mathbf{\Sigma}_{e}$, 
 $n_a$, $n_d$ and $n_S$ for identifying the pure sources. $\mathbf{\Sigma}_{e}$ is error variance-covariance matrix assumed to be a diagonal matrix.\par
It is worthwhile to notice that in the process of estimating $\mathbf{\Sigma}_e$ through DIPCA, the identifiability constraint discussed in \cite{narasimhan2008model} and \cite{maurya2018identification} never gets violated. This is because there exists at least a lag $L \geq \eta_p$, such that $d_L(d_L+1) \geq 2M(L+1)$. \par

Initially, the noise variance-covariance matrix $\mathbf{\Sigma}_e=\text{diag}(\sigma^2_{e_i})$ is estimated using DIPCA at an arbitrary large lag $L$. Subsequently, the data matrix $\mathbf{Z}$ is scaled with $\mathbf{\Sigma}_e^{-1/2}$ to obtain $\mathbf{Z}_s$. Now, to determine $n_a$, DIPCA is applied to the variable data matrix $\mathbf{Z}_s$ without any lag $L$, i.e. $L=0$. The number of unity singular values equals the value of $n_a$.
    The data matrices of all the scaled variables 
   at two arbitrary large lags, $L_1$ and $L_2$ are subjected to DIPCA. The corresponding number of unity singular values are $d_{L_1}=d_1$ and $d_{L_2}=d_2$, respectively. This information is used to calculate $n_d$ from (\ref{expression for nd}) 
 \begin{align}
     n_{d_1}+n_{d_2}\ldots=\frac{d_1-d_2}{L_1-L_2}-n_a, \\
     n_d=\frac{d_1-d_2}{L_1-L_2}-n_a
     \label{expression for nd},
 \end{align} 
 where equation (\ref{expression for nd}) is derived from (\ref{equation relating lag, order and number of equations}) and (\ref{sum of number of dynamical equations}). Using $n_a$ and $n_d$, we get $n_S=M-n_a-n_d$.\par
 
  To determine the free variables, the entire set of variables is divided into two subsets of cardinalities $n_a+n_d$ and $M-n_a-n_d$. The condition number of the coefficient matrix of $n_a+n_d$ variables' instantaneous terms, assumed to be the set of dependent variables i.e. $\mathbf{A}_\text{dep}$ is evaluated. This procedure is repeated for all the combinations in $\binom{M}{n_a+n_d}$. The coefficient matrix is derived from the corresponding right singular vectors of unity singular values from the above step of DIPCA with a user-specified maximum lag $L$. Instead of determining the rank from rotated constraints, condition numbers are employed, as noise ensures rank fulfilment of the coefficient matrix of variables.  Now, arranging the combinations in the order of decreasing condition number, combinations with a condition number less than the threshold of $10$ are admissible for pure sources determination. This threshold has been put heuristically, determined from Monte-Carlo simulations, and further investigations must be done.\par
 
 An admissible combination gives a dependent set of variables and free variables (pure sources). Also, as discussed earlier, the count of pure sources appears to be greater than or equal to $n_S$ in DAE systems. The entire methodology of the above algorithm is summarized in Table \ref{tab: table 1}.
 \begin{table}[h]
     \centering
     \caption{Algorithm to determine the pure sources using PoV method}
     \begin{tabular}{p{3.3in}}
     \hline
         \begin{enumerate}
         \item Estimate the error variances ($\sigma_{e_i}^2$) of the variables.
         \begin{itemize}
            \item Apply DIPCA on all the variables stacked with an arbitrary maximum lag ($L$).
            \item Pick up the error variances of the variables after DIPCA converges.
        \end{itemize}
          \item Scale the variables with their corresponding error standard deviations ($\sigma_{e_i}$).
         \item Determine the number of algebraic constraints ($n_a$).
         \begin{itemize}
             \item Apply DIPCA on the scaled variables data matrix with a lag of $0$.
             \item Count the number of unity singular values.
         \end{itemize}
         \item Determine the number of dynamical constraints ($n_d$).
         \begin{itemize}
             \item Apply DIPCA on the data matrix, stacking all the scaled variables with a user-specified maximum lag.
             \item Count the number of unity singular values at two different lags and use (\ref{expression for nd}) to get $n_d$.
         \end{itemize}
         \item Calculate the number of pure sources $n_S = M - (n_a + n_d)$, where $M$ is the number of variables.
        \item Calculate the condition number of all the combinations.
        \begin{itemize}
            \item Partition the variables into two subsets of cardinalities $n_a+n_d$ and $M-n_a-n_d$.
            \item Form the coefficient matrix $\mathbf{A}_\text{dep}$ of $n_a + n_d$ variables' instantaneous terms from rotated constraints obtained in step $4$:\begin{align*}
            \mathbf{A}_{\text{dep}}\mathbf{x}_{\text{dependent}} =-\mathbf{A}_{\text{r}}\mathbf{x}_{\text{rest}}
            \end{align*}

            \item Calculate the condition number of $\mathbf{A}_{\text{dep}}$ from each combination. 
            \item Arrange the combinations in the descending order of condition numbers.
        \end{itemize}
        \item Finally, obtain the pure sources.
        \begin{itemize}
            \item Select the combinations with a condition number less than the threshold as admissible combinations.
            \item Pick up the free variables from the admissible combinations and form the pure sources set.
            \item Classify the sources into unambiguous and ambiguous.
        \end{itemize}
          \end{enumerate}\\
     \hline
     \end{tabular}
     \label{tab: table 1}
 \end{table}

\section{Results and Discussion}\label{sec:case_studies}
Few case studies have been demonstrated to showcase the efficacy of the proposed method in identifying the pure causal variables in the LTI-DAE system, especially using the measurements of the variables assumed to be contaminated with zero mean Gaussian white noise.\par
The first case study explains the applicability of this method when $n_a = 0$. 
The second case study deals with pure sources in the feedback of a system.
\subsection*{Case study I}\label{case study I}
Consider a two-tank liquid level system governed by solely difference equations, as follows:
\begin{align}
\begin{array}{l}
    h_1[k] = 0.659h_1[k-1] + 0.6815F_0[k-1], \\
    h_2[k] = 0.7168h_2[k-1] + 0.1772h_1[k-1] 
    \end{array} \label{liquid level system governing equations},
\end{align}
where $h_1$ and $h_2$ are the liquid levels of the tanks connected in series, constituting a non-interacting system. This designation of non-interacting arises from the fact that variations in $h_2$ do not affect the transients in $h_1$. Whereas $F_0$ is the input flow rate to the tank one. Expressing (\ref{liquid level system governing equations}) in the form of a constraint matrix, as (\ref{eq:dae_sys}), yields:

\begin{align}
\begin{bmatrix}
    0&1&0&-0.6815&-0.659&0\\
    0&0&1&0&-0.1772&0.7168
\end{bmatrix}\begin{bmatrix}
    F_0[k]\\h_1[k]\\h_2[k]\\F_0[k-1]\\h_1[k-1]\\h_2[k-1]
\end{bmatrix}=\mathbf{0}  \label{constraint form of equations in case study I}.  
\end{align}

Only the entire first column of $\mathbf{A}$ in (\ref{constraint form of equations in case study I}) has zeros corresponding to $F_0[k]$ coefficient, implying that $F_0$ is the unambiguous pure source.\par
\begin{figure}[h]
    \centering
    \begin{tikzpicture}[node distance={13mm},thick,main/.style = {draw, circle}]
\node[main](1){$F_0$};
\node[main](2)[below left of=1]{$h_1$};
\node[main](3)[below right of=1]{$h_2$};
\draw[->] (1) -- (2);
\draw[->] (2) -- (3);
\end{tikzpicture}
    \caption{The true network topology of the two-tank liquid level system.}
    \label{two tank system network topology}
\end{figure}
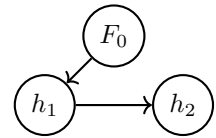

 
With $M=3$, $n_a=0$, and $n_d=2$, it follows that $n_S=M-n_a-n_d=1$. Since $n_a=0$, there are no ambiguous sources in the system. The ground truth is that the pure source variable is $F_0$. In Fig. \ref{two tank system network topology},
observe, $F_0$ affects both $h_1$ and $h_2$ but remains unaffected by any other variable. The directed edge between two nodes indicates that
there is a difference equation that governs the relationship between the two variables \par

However, none of the aforementioned information is available; solely noisy measurements are accessible for analysis and the labels are concealed. Specifically, $ z_1,z_2,$ and $z_3$ are provided, where $z_1 = h_1+e_1$, $
z_2 = h_2+e_2,$ and $z_3 = F_0+e_3$, with $ e_1,e_2,e_3$ representing the measurement errors. During the simulation, SNRs of $z_1$, $z_2$, and $z_3$ were set to $2$, $3$, and $5$, respectively. Correspondingly, the true error variances used were $\sigma_{e_1}^2 = 0.41$, $\sigma_{e_2}^2 = 0.049$, and $\sigma_{e_3}^2 = 0.2$. From DIPCA, the estimated error variances in $z_1$, $z_2$, and $z_3$  were $\hat{\sigma}^2_{e_1}=0.427$, $\hat{\sigma}^2_{e_2}=0.051$, and $\hat{\sigma}^2_{e_3}=0.181$, respectively. The variables were then scaled using their respective estimated error standard deviations to proceed with the algorithm. No unity singular values were observed when applying DIPCA to the scaled variables' data matrix, resulting in $n_a = 0$. 
Subsequently, when DIPCA was applied on the scaled variables' data matrices stacked with lags of $L = 5$ and $ L = 1$, resulted in the observation of $d_1 = 10$ and $d_2 = 2$ unity singular values, respectively. Utilizing (\ref{expression for nd}), with these two lags, $n_d = 2$ was obtained. This yielded two difference equations and zero algebraic equations, signifying the existence of two dependent variables and one pure source $(n_S = M - n_a - n_d = 3 - 0 - 2 = 1)$.\par

From the step of DIPCA with a maximum lag of $5$, corresponding to unity singular values, resulted in the rotated constraint matrix $\mathbf{TA}$ from the right singular vectors, where $\mathbf{T}$ is an unknown similarity transformation. The complete set of variables was then partitioned into dependent and remaining variables. The condition numbers of the coefficient matrix for the instantaneous terms of the two dependent variables, as described in (\ref{partition of variables}), were calculated for all possible combinations based on the rotated constraints. The resulting condition numbers were arranged in descending order and presented in Table \ref{table combinations - case study I}. Upon analysis, it was observed that the third combination exhibited a condition number less than $10$, making it suitable for determining the pure source. Consequently, $z_3$ was identified as an `unambiguous' pure source commensurate with ground truth.\par

The findings of type I and type II errors using the proposed approach were obtained from $100$ realizations, with $N = 2047$ observations, and at different arbitrary low SNRs are presented in Table \ref{type-I and type II errors of case study I at SNR 2,3,5} and \ref{type-I and type II errors of case study I at SNR 2,1.5,1}. Type I errors represent the percentage of cases of that combination with a condition number which showed false positives using the proposed threshold. Similarly, type II errors represent the percentage of cases that showed false negatives.\par

\begin{table}[h]
    \centering
    \caption{All combinations showing the admissibility $\alpha$ from their condition number in case study I}
    \begin{tabular}{|C{2.5cm}|C{2.5cm}|c|c|c|c|}
    \hline
   Candidate dependent variables&  Condition number of $\mathbf{A}_\text{dep}$ &  $\alpha$& Free variable\\
    \hline
       \{$z_1[k],z_3[k]$\}& $201.59$ & $0$&-\\
       \{$z_2[k],z_3[k]$\} & $69.78$ & $0$&-\\
       \{$z_1[k],z_2[k]$\} & $1.229$ & $1$& $z_3$\\
        \hline
    \end{tabular}
    \label{table combinations - case study I}
\end{table}
\begin{table}[h]
    \centering
    \caption{Type I and Type II errors of the case study I where
    SNRs of $z_1$, $z_2$ and $z_3$ were set to $2$, $3$, and $5$, respectively}
    \begin{tabular}{|C{2.1cm}|C{2.1cm}|C{2.1cm}|}
    \hline
        Combination & Type I errors (\%) & Type II errors (\%) \\
        \hline
         $\{z_3,z_1\}$&$0$&-\\
         
         $\{z_3,z_2\}$&$0$&-\\
         
         $\{z_1,z_2\}$&-&$1$\\
         \hline
    \end{tabular}
    \label{type-I and type II errors of case study I at SNR 2,3,5}
\end{table}

\begin{table}[h]
    \centering
    \caption{Type I and Type II errors of the case study I where
    SNRs of $z_1$, $z_2$ and $z_3$ were set to $2$, $1.5$, and $1$, respectively}
    \begin{tabular}{|C{2.1cm}|C{2.1cm}|C{2.1cm}|}
    \hline
        Combination & Type I errors (\%) & Type II errors (\%) \\
        \hline
         $\{z_3,z_1\}$&$3$&-\\
         
         $\{z_3,z_2\}$&$2$&-\\
         
         $\{z_1,z_2\}$&-&$1$\\
         \hline
    \end{tabular}
    \label{type-I and type II errors of case study I at SNR 2,1.5,1}
\end{table}
Table \ref{type-I and type II errors of case study I at SNR 2,1.5,1} shows 
Type I and Type II errors, with a commonly accepted significance level of $5\%$ and likely due to the low SNR.

 \subsection*{Case study II}\label{case study II}
The peculiarity of examining this particular case from \cite{VishweshRamanathan} lies in the existence of signal feedback, wherein the pure signal without any measurement error gets fed back into the system. The challenge of estimating the error variances when the measurement error propagates through another variable in feedback has not yet been solved by DIPCA. Specifically, the estimation of the non-diagonal error variance-covariance matrix with correlated errors through DIPCA remains unresolved. 
The models estimated in this work from DIPCA with a strict assumption of no measurement error feedback fall under the category of \textit{output-error} (OE) models. One such case is the study in an RC circuit, as the measurements of voltages, current, and capacitance are not fed back into the system.\par

An RC circuit connected to a voltage source $U$, where the current $i$ flowing through the circuit induces voltage drops across the resistor$({R})$ is $V$, and the capacitor$({C})$ is $X$. 
\begin{figure}[h]
    \centering
    \begin{circuitikz}[american, scale = 1.5][americanvoltages]
  \draw (0,0)
  to[V_=$U$] (0,-2) 
  to [short = $i$](2,-2)  ;
  \draw
  (0,0) to[R, l^=${R}$, v=$V$] (2,0) 
    to[C, l^=${C}$, v = $X$] (2,-2) 
  ;
\end{circuitikz}
    \caption{RC circuit.}
    \label{RC circuit}
\end{figure}
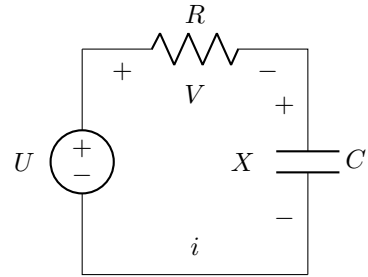
 Ohm's law describes the fundamental relationship governing the voltage drop across the resistor:
\begin{align}
V[k]=R\hspace{0.1cm}i[k].
\end{align}
Similarly, the voltage across the capacitor is influenced by the time-dependent accumulation of charge and its evolution in discrete time is given by the formula:
\begin{align}
X[k+1]=e^{-\Delta t/RC}X[k]-(e^{-\Delta t/RC}-1)U[k].
\end{align}
On substituting the values $R=50$, $C=1$, and $\Delta t =1$, we get the governing equations, 
\begin{align}
\begin{array}{l}
        X[k+1] =0.9802X[k]+0.0198U[k],\\
         U[k] = X[k]+V[k],\\
        i[k] = 0.02V[k].
\end{array}\label{RC circuit governing equations}
\end{align}
Writing them in the constraint matrix form $\mathbf{Ax}[k,\eta]=\mathbf{0}$ 

\begin{align}
    \begin{bmatrix}
        0&0&1&0&\boxtimes&\ldots\\
        1&-1&-1&0&\boxtimes&\ldots\\
        0&-0.02&0&1&\boxtimes&\ldots
    \end{bmatrix} 
    \begin{bmatrix}
        U[k]\\V[k]\\X[k]\\i[k]\\U[k-1]\\\vdots\\i[k-1]
    \end{bmatrix}=\mathbf{0}
\end{align}
where the elements $\boxtimes$ are of no interest. With $n_S = M-n_a-n_d = 4-2-1 = 1$ and no columns of zeros for instantaneous terms, it can be inferred that no unambiguous pure source variables exist. However, it is noted that the number of sources seems to exceed a count of one due to its integration within the algebraic equations. The same is observed from the table {\ref{tab: RC circuit table of deterministic system}}. The ground truth about the pure source is that it can be either $U$, $V$, or $i$. \par
\begin{table}
    \centering
    \caption{Combinations showing the pure source candidates in the RC circuit system from the governing equations}
    \begin{tabular}{|c|c|c|}
    \hline
    Candidate dependent variables&Rank of $\mathbf{A}_\text{dep}$&Free variables\\
    \hline
         \{$U[k]$, $V[k]$, $i[k]$\}& $2$& -\\
         \{$U[k]$, $V[k]$, $X[k]$\}& $3$ & $i$\\
         \{$V[k]$, $X[k]$, $i[k]$\}& $3$& $U$\\
         \{$U[k]$, $X[k]$, $i[k]$\}& $3$& $V$\\
         \hline
    \end{tabular}
    \label{tab: RC circuit table of deterministic system}
\end{table}

Now, when given alone the measurements of the variables  $z_1= U+e_1, z_2 =V+e_2, z_3 = X+e_3, z_4 = i+e_4$ unlabelled, identifying the pure source among them is the problem, where $\{e_1,e_2,e_3,e_4\}$ are zero-mean Gaussian white noises of unknown variance. \par
During simulation, the SNRs of $z_1$, $z_2$, $z_3$ and $z_4$ were set to $5$, $4$, $6$, and $8$, respectively. Initially, the variables' error variances were estimated using DIPCA, and the variables were scaled accordingly. The estimated error variances in $z_1$, $z_2$, $z_3$ and $z_4$ were $9.151$, $12.305$, $0.0417$, and $0.0483$, respectively. Following this, the number of algebraic equations was estimated. At lag $L=0$, the singular values obtained were $ 5.31$, $4.99$, $1.02$, and $0.98$. As two unity singular values were observed, it follows that $n_a=2$. Similarly,  at lag $L=2$, the singular values were $8.92$, $5.24$, $5.05$, $4.95$, $1.07$, $1.05$, $1.03$, $0.99$, $0.98$, $0.96$, $0.94$, and $0.93$ among which there were $8$ unity singular values. Using (\ref{expression for nd}),  $n_d$ was equal to one. Therefore, the number of sources was $n_S=4-2-1=1$. Using the rotated constraints that were obtained from the step of calculating $n_d$, the condition numbers of the $\mathbf{A}_\text{dep}$ of all the combinations were calculated and shown in table \ref{tab: Data-driven PoV approach of case study IV}.

\begin{table}[]
    \centering
    \caption{Combinations showing admissibility along with the free variables in case study II}
    \begin{tabular}{|>{\centering}p{2.4cm}|>{\centering}p{2.4cm}|c|c|}
    \hline
    Candidate dependent variables& Condition number of $\mathbf{A}_\text{dep}$ & $\alpha$ & Free variables\\
    \hline
    \{$z_1$, $z_2$, $z_4$\} &$35.123$ &$0$ & -\\
        \{$z_1$, $z_2$, $z_3$\} & $4.8651$  & $1$&$z_4$ \\
        \{$z_1$, $z_3$, $z_4$\} &$1.57$ & $1$& $z_2$\\
        \{$z_2$, $z_3$, $z_4$\} &$1.34$ &$1$ & $z_1$\\
         \hline
    \end{tabular}
    
    \label{tab: Data-driven PoV approach of case study IV}
\end{table}

The pure source variable candidates were either $z_1$, $z_2$ or $z_4$, commensurate with the ground truth. With $100$ Monte-Carlo simulation runs, $N = 1000$ observations, and at different SNRs, the Type I and Type II errors obtained are presented in table \ref{type-I and type II errors of case study II at SNR 5,4,6,8} and in table \ref{type-I and type II errors of case study II at SNR 6,4,5,5}. Even in this case, these errors are likely due to low SNRs.
\begin{table}[h!]
        \centering
        \caption{Type I and Type II errors of case study II where SNRS of $z_1$, $z_2$, $z_3$, and $z_4$ were set to 5, 4, 6, and 8, respectively}
        \begin{tabular}{|c|c|c|}
        \hline
        Combination & Type I error (\%) & Type II error (\%) \\
        \hline
        $\{z_1,z_2,z_4\}$ &0&-\\
        
        $\{z_1,z_2,z_3\}$  & -&9\\
        
        $\{z_2,z_3,z_4\}$  &-&3\\
        
        $\{z_1,z_3,z_4\}$ &-&6\\
        \hline
        \end{tabular}
        \label{type-I and type II errors of case study II at SNR 5,4,6,8}
    \end{table}

\begin{table}[h!]
        \centering
        \caption{Type I and Type II errors of case study II where SNRS of $z_1$, $z_2$, $z_3$, and $z_4$ were set to 6, 4, 5, and 5, respectively}
        \begin{tabular}{|c|c|c|}
        \hline
        Combination & Type I error (\%) & Type II error (\%) \\
        \hline
        $\{z_1,z_2,z_4\}$ &0&-\\
        
        $\{z_1,z_2,z_3\}$  & -&13\\
        
        $\{z_2,z_3,z_4\}$  &-&7\\
        
        $\{z_1,z_3,z_4\}$ &-&6\\
        \hline
        \end{tabular}
        \label{type-I and type II errors of case study II at SNR 6,4,5,5}
    \end{table}

\section{Conclusions \label{sec:conclusions}}

In this work, ambiguous and unambiguous pure sources within the error-in-variables (EIV) framework has been identified. Pure sources, part of the sole difference equations, are inherently unambiguous. Our investigation has elucidated that when the pure sources are integrated within algebraic equations, there appears to be an excess in the number of pure sources due to instantaneous relationships among the variables. These ambiguous sources are called potential candidates of pure sources.\par

The PoV method has been employed to identify the pure sources, regardless of the equation types (only difference equations or a mix of algebraic and difference) within the system. Therefore, this PoV is superior to the method that was presented in \cite{kathari2022novel}, as PoV can also determine pure sources in dynamical systems having no algebraic equations. Practically, when confronted with measurements of variables with errors, the DIPCA combining with PoV proved to be an effective tool in identifying these sources.\par

Notably, temporal precedence information like delay between the variables as shown in \cite{kathari2022novel} has not been used in this work to discern pure sources. As variables are sometimes related instantaneously, temporal precedence information is rendered unsuitable for our purposes. Furthermore, our approach demonstrated that identifying the `exact' governing multi-input multi-output differential-algebraic equations (MIMO-DAE) is unnecessary i.e., the constraint matrix ($\mathbf{A}$) tying up all the variables; solely the rotated constraints ($\mathbf{TA}$) suffice for the identification of sources.\par

The scope of this work applies to open-loop and limited to closed-loop systems where the signals are measured with errors, but no measurement errors are fed back into the system. These classes of models are called OE models. Moreover, the measurement error is assumed to be Gaussian white noise. A future direction is identifying the sources in LTI-DAE systems corrupted with colored noise. An important limitation of this work is the ambiguity regarding the potential candidates for sources in OE models, which is yet to be clarified. Few other future directions of research are - causal discovery in LTI-DAE systems within classical model framework as some variables may not be corrupted with measurement errors and in multi-scale LTI-DAE systems.

\section*{ACKNOWLEDGMENT}


We would like to express our sincere gratitude to Professor Shankar Narasimhan, IIT Madras for generously providing the DIPCA code, which played an integral role in every step of the algorithm proposed in this paper. 

\end{document}